\documentclass[journal]{IEEEtran}

\usepackage{amsmath,amssymb,amsfonts}
\usepackage{graphicx}
\usepackage{algorithm,algorithmic}
\usepackage{booktabs}

\usepackage{cite} 
\usepackage[colorlinks=true, linkcolor=blue, citecolor=blue, urlcolor=blue]{hyperref}

\makeatletter
\def\@citex[#1]#2{\leavevmode
  \let\@citea\@empty
  \@cite{\@for\@citeb:=#2\do
    {\@citea\def\@citea{,\penalty\@m\ }%
     \edef\@citeb{\expandafter\@firstofone\@citeb\@empty}%
     \if@filesw\immediate\write\@auxout{\string\citation{\@citeb}}\fi
     \@ifundefined{b@\@citeb}{\hbox{\reset@font\bfseries ?}%
       \G@refundefinedtrue
       \@latex@warning
         {Citation `\@citeb' on page \thepage \space undefined}}%
       {\Hy@raisedlink{\hyper@anchorstart{cite.\@citeb}\hyper@anchorend}%
        \hyper@linkstart{cite}{cite.\@citeb}%
        \@cite@ofmt{\csname b@\@citeb\endcsname}%
        \hyper@linkend}}}{#1}}
\makeatother

\usepackage[capitalize,nameinlink]{cleveref}

\crefname{section}{Sec.}{Secs.}
\Crefname{section}{Section}{Sections}
\crefname{table}{Tab.}{Tabs.}
\Crefname{table}{Table}{Tables}
\crefname{figure}{Fig.}{Figs.}
\Crefname{figure}{Figure}{Figures}

\def\BibTeX{{\rm B\kern-.05em{\sc i\kern-.025em b}\kern-.08em
    T\kern-.1667em\lower.7ex\hbox{E}\kern-.125emX}}

\begin{document}

\title{3D Ultrasound-Derived Pseudo-CT Synthesis Using a Transformer-Augmented Residual Network for Real-Time Operator Guidance}
\author{Sapna Sachan\thanks{Sapna Sachan is with Indian Institute of Technology Guwahati, Assam, PIN 781039 India (e-mail: s.sapna@iitg.ac.in)}, and Amulya Kumar Mahto \thanks{Amulya Kumar Mahto is with Indian Institute of Technology Guwahati, Assam, PIN 781039 India, (email: akmahto@iitg.ac.in)}}

\maketitle

\begin{abstract}
Computed tomography (CT) is indispensable for clinical diagnosis and image-guided interventions but exposes patients to ionizing radiation, motivating the development of safer imaging alternatives. Ultrasound (US) is non-ionizing and widely accessible; however, it is highly operator dependent and lacks quantitative tissue characterization, often leading to diagnostic uncertainty and unnecessary CT examinations. This work presents a 3D ultrasound-derived pseudo-CT (UD-pCT) framework that generates CT-like anatomical reference volumes inferred from US, without aiming to reproduce physically accurate Hounsfield Units. Paired 3D kidney US and CT volumes from the TRUSTED dataset are first spatially aligned using a landmark-based multimodal registration pipeline, creating high-quality paired inputs for supervised training of an adversarial framework. The proposed Bottleneck Transformer Residual U-Net3D (BT-ResUNet3D) model employs a 3D residual encoder–decoder generator augmented with a transformer bottleneck, enabling effective modeling of fine-grained local anatomical structures as well as long-range volumetric dependencies, while a 3D Conditional PatchGAN discriminator enforces local structural realism in the synthesized pseudo-CT volumes. Quantitative evaluation using PSNR and SSIM demonstrates that the proposed method outperforms established baselines in structural fidelity and perceptual image quality. The UD-pCT volumes provide real-time anatomical reference for operator guidance, potentially reducing acquisition variability and unnecessary CT use. A limitation of this study is the relatively small paired dataset, which may limit the generalizability of the proposed model.
All data preprocessing and testing codes are given as supplimentary material to ensure the reproducibility of the experimental results.
\end{abstract}


\section{Introduction}
\label{sec:introduction}
\IEEEPARstart{I}{maging} with computed tomography (CT) is a cornerstone of modern clinical practice, supporting diagnosis, treatment planning, and image-guided interventions. Despite its clinical utility, CT imaging relies on ionizing radiation, which is associated with a small but measurable risk of radiation-induced cancer for each scan. With an estimated 375 million CT examinations performed globally each year~\cite{bib5,bib6}, this risk becomes significant at the population level. At the same time, the global burden of cancer continues to rise, with approximately 20 million new cases reported in 2022 and projections reaching nearly 35 million by 2050~\cite{bib1,bib2}. Earlier studies estimated that almost 29,000 future cancer cases in the United States could be linked to CT scans performed in 2007 alone~\cite{bib5}. More recent analyses suggest that, if current imaging practices continue, CT-associated cancers could account for up to $\sim$5\% of among all newly diagnosed cases~\cite{bib4}.
These trends emphasize the urgency of minimizing CT examinations which can be avoided, particularly in clinical cases where some safe imaging modalities can give sufficient anatomical guidance. This motivates the development of some useful complementary approaches that reduce radiation exposure and also preserve the clinical utility of image-guided workflows.

US imaging is a safe, non-ionizing, portable, and cost-effective alternative~\cite{bib7}. However, US has its own inherent limitations: its intensity values do not correspond directly to tissue electron density or Hounsfield Units (HU), which restricts quantitative interpretation, and quality of an image is highly dependent on the operator of the instrument. In particular, image quality gets significantly affected by the varying probe pressure applied by the operator, which again leads to incomplete anatomical visualization, potential misinterpretation, and, consequently, leading to unnecessary follow-up CT examinations.

A encouraging strategy to address these limitations is by bridging the modality gap by generating 3D UD-pCT volumes using US data. It is important to note here that the goal of UD-pCT synthesis is not to replace diagnostic CT imaging, but rather to provide real-time anatomical reference information during US acquisition. By presenting a CT-like anatomical context, UD-pCT volumes can enhance effectiveness of operator by increasing the awareness of probe positioning, imaging orientation, and anatomical coverage. Such guidance can help steer acquisitions toward optimal imaging planes, improving consistency of imaging among operators, and diminishing unnecessary CT referrals, while fully preserving the safety and accessibility advantages of US.

However, existing deep learning models such as 3D ResUNet-based models, while effective at local feature extraction, rely on convolutional receptive fields and struggle to capture global anatomical relationships across large 3D volumes. ResViT-style architectures, which replace or heavily integrate transformers throughout the encoder–decoder pathway, improve global context modeling but often incur high computational cost and may degrade fine-grained local anatomical detail—particularly problematic for noisy US data.

Motivated by these developments, this work proposes BT-ResUNet3D model for UD-pCT volumes synthesis using paired kidney US and CT volumes from the TRUSTED dataset. Unlike ResViT models, our approach introduces a lightweight transformer exclusively at the bottleneck, where it models long-range volumetric dependencies without disrupting early-stage convolutional feature extraction. In contrast to standard 3D ResUNet, deep residual blocks are combined with global transformer context through residual fusion, enabling the network to preserve local anatomical fidelity while enhancing global structural coherence. The resulting UD-pCT volumes are designed to function as real-time anatomical references for operator guidance, rather than diagnostic CT substitutes.

The main contributions of this work are summarized as follows:
\begin{itemize}
\item We propose a robust and clinically motivated \textbf{3D multimodal data preprocessing pipeline}, consisting of landmark-based US–CT registration followed by region-of-interest (ROI) cropping and field of view masking, to construct anatomically consistent and spatially aligned volumetric training pairs, which is critical for reliable supervised cross-modality synthesis.
\item We introduce a transformer-augmented bottleneck integrated within a 3D residual encoder--decoder(BT-ResUNet3D) architecture to explicitly capture long-range volumetric dependencies beyond convolutional receptive fields, while preserving local anatomical fidelity.
\item We incorporate a \textbf{3D conditional PatchGAN discriminator} to enforce local structural realism and enhance high-frequency anatomical details in the synthesized UD-pCT volumes.
\item We stablish the clinical utility of UD-pCT volumes for operator guidance for US acquisition, with the ability to improve scan consistency and reduced unnecessary CT usage.
\item Quantitative and qualitative evaluation on the TRUSTED dataset shows that the proposed method outperforms baseline approaches on comparing in terms of PSNR and SSIM, leading to improved structural fidelity and perceptual quality.
\end{itemize}

\section{Literature Survey}

\subsection{Operator Guidance in Ultrasound Imaging}
US imaging is highly operator dependent, and image quality strongly fluctuates with probe orientation, pressure, and anatomical window selection. Several works address these important challenges, focusing on variety of strategies to enhance quality of US imaging through artificial intelligence and multi-modal integration to provide real-time guidance to operators, which improves acquisition consistency and diagnostic reliability. Considering these issues, Ossaba et al.~\cite{bib13} validated a modular deep learning prototype that assists operator in acquiring diagnostic US images, for improved quality of captured planes. Similarly, Baloescu et al.~\cite{bib14} demonstrated that AI-assisted lung US helps non-experts to achieve performance comparable to trained clinicians. Recent reviews highlight that AI-guided acquisition, standard plane detection, and semantic segmentation are some of the critical components for reducing operator variability~\cite{bib15}. Hybrid approaches that integrate data-driven learning with physics-informed constraints further improve anatomical fidelity and acoustic consistency in US-based modeling~\cite{bib11}. Even though, deep regression model ~\cite{xing2025deep} shows promise in correcting for liver motion and improving anatomical alignment, fully mimicking expert-level adaptability across diverse clinical scenarios remains challenging.

\subsection{US–CT Registration Methods}
Multimodal registration between US and CT is essential for cross‑modality synthesis and operator guidance. Classical intensity‑based registration methods, namely normalized cross‑correlation (NCC) and correlation ratio (CR), are used to align misaligned CT and US images by optimizing similarity metrics despite differing imaging physics~\cite{bib18,bib19}. Landmark‑based automatic registration approaches avoid manual initialization by localizing anatomical fiducials in both modalities and using them to compute rigid or similarity transforms, achieving robust alignment without direct intensity dependency~\cite{bib17}. Hybrid deep learning‑assisted methods combine both segmentation and feature extraction with conventional similarity metrics to guide 3D CT US registration, improving robustness by focusing on meaningful anatomical structures~\cite{bib16}.

\subsection{US-to-CT Synthesis and Cross-Modality Generation}

Recent advances in deep learning have enabled powerful nonlinear mappings between heterogeneous imaging modalities. Convolutional neural networks (CNNs) have been widely adopted for medical image synthesis due to their strong capacity for local representation learning~\cite{bib8}. Generative adversarial networks (GANs) support both paired and unpaired image translation: conditional GAN frameworks such as pix2pix enable supervised paired synthesis~\cite{bib20}, while CycleGAN variants facilitate unpaired learning when paired data are limited. Transformer-based architectures have demonstrated strong performance in representation learning and synthesis by explicitly modeling long-range dependencies~\cite{bib10}, which are critical for maintaining global anatomical consistency in 3D volumes. Methods such as Double U‑Net CycleGAN further improve spatial coherence in volumetric data by leveraging adjacent slice information~\cite{bib21}.

Classical volumetric architectures remain foundational: 3D U‑Net captures hierarchical spatial features in 3D data~\cite{bib31}, and 3D ResNet variants stabilize deep learning via residual connections across volumetric layers~\cite{bib32}. Residual vision transformer approaches (e.g., ResViT) integrate transformer blocks with convolutional representations throughout the network, enhancing global context modeling at the cost of increased computational complexity~\cite{bib33}.
    
More recently, diffusion-based generative models have shown improved stability and anatomical fidelity for 3D medical synthesis. Denoising diffusion probabilistic models provide a principled generative framework with strong performance on high-resolution volumetric data~\cite{bib34}, and latent diffusion approaches enable efficient high-quality generation by operating in a learned latent space~\cite{bib35}. Conditional wavelet diffusion models have been proposed to further improve cross-modality translation quality by capturing multi-scale anatomical features~\cite{bib36}, while cross-conditioned diffusion designs enhance robustness in heterogeneous modality translation tasks~\cite{bib37}.

Despite these advances, US-to-CT synthesis remains challenging due to modality-specific noise, acoustic shadowing, and limited availability of paired data. GAN-based pix2pix models show feasibility in controlled settings but often generalize poorly to real clinical US data. Unpaired CycleGAN variants improve structure consistency but lack voxel-level supervision and struggle with global 3D context modeling, resulting in ambiguous intensity mapping and reduced anatomical fidelity. These limitations motivate the development of volumetric architectures that jointly capture fine-grained local textures, preserve anatomical structures through deep residual learning, and model long-range spatial dependencies across 3D volumes.

In contrast to ResViT and multi-scale transformer architectures, which embed self-attention mechanisms throughout the convolutional feature hierarchy, the proposed method introduces a lightweight transformer module exclusively at the network bottleneck. This design enables the capture of long-range 3D contextual dependencies while preserving local anatomical detail through convolutional feature extraction, with global and local representations fused via residual addition. Unlike prior approaches that often assume pre-aligned inputs, our framework is trained on robustly registered and spatially standardized paired US and CT volumes obtained through landmark-based multimodal registration and region-of-interest cropping. This preprocessing strategy ensures accurate voxel-wise correspondence between modalities, which is critical for reliable cross-modality learning. As a result, the proposed BT-ResUNet3D model synthesizes anatomically consistent pseudo-CT volumes from US, supporting operator guidance during US acquisition rather than diagnostic CT replacement

\section{Dataset Description}

This work uses the publicly available TRUSTED dataset ~\cite{bib30}, which provides paired three-dimensional computed tomography (CT) and transabdominal US data for kidney-focused segmentation, registration, and cross-modality analysis. The dataset includes expert annotations and standardized evaluation protocols to support supervised learning and reproducible experimentation.
The CT subset contains 48 volumetric scans, each comprising both left and right kidneys (96 kidneys total). For each CT volume, kidney segmentation masks annotated by two independent experts and an estimated ground-truth mask are provided, along with anatomical landmarks, kidney surface meshes in image and CT device coordinate systems, transformation matrices, and predefined five-fold patient-level cross-validation splits.
The US subset consists of 59 reconstructed 3D volumes, each corresponding to a single kidney, acquired using tracked freehand transabdominal US under realistic clinical conditions. Each US volume includes expert kidney segmentation masks, estimated ground truth, anatomical landmarks, and kidney surface meshes in both image and US device coordinate systems. Five-fold kidney-level cross-validation splits are defined, explicitly separating left and right kidneys across patients. These paired data and annotations are leveraged for 3D US–CT registration, aligned training pair construction, and evaluation of the proposed UD-pCT synthesis framework.

\section{Data Preprocessing}
\label{sec:preprocessing}

To construct anatomically consistent and spatially aligned training pairs for supervised 3D US–CT synthesis, a multi-stage preprocessing pipeline was employed. The pipeline comprises (i) landmark-based multimodal registration, (ii) region-of-interest (ROI) cropping and (iii) field of view masking, with each step progressively reducing geometric discrepancies between modalities and ensuring the voxel-wise correspondence required for learning reliable cross-modality mappings. Fig \ref{fig:registered} shows both the CT and US volumes before and after the pre-processing steps.

\subsection{Landmark-Based CT–US Registration}

Due to differences in patient positioning, imaging geometry, and acquisition protocols, CT and US volumes are not inherently aligned. To achieve spatial correspondence between modalities, a landmark-based similarity registration was performed using paired anatomical landmarks.

Let $\mathbf{P} = \{\mathbf{p}_i\}_{i=1}^{N}$ denote US landmarks and $\mathbf{Q} = \{\mathbf{q}_i\}_{i=1}^{N}$ denote the corresponding CT landmarks, both expressed in world coordinates. The goal is to estimate a similarity transformation consisting of a rotation matrix $\mathbf{R} \in SO(3)$, an isotropic scaling factor $s$, and a translation vector $\mathbf{t}$ such that the transformed US landmarks best align with the CT landmarks. This is achieved by minimizing
\begin{equation}
\min_{s,\mathbf{R},\mathbf{t}} \sum_{i=1}^{N} \left\| \mathbf{q}_i - (s \mathbf{R} \mathbf{p}_i + \mathbf{t}) \right\|^2.
\end{equation}
The optimization is solved using the Umeyama algorithm, which provides a closed-form solution based on singular value decomposition. The resulting transformation is represented in homogeneous form as
\begin{equation}
\mathbf{T}_{\text{US}\rightarrow\text{CT}} =
\begin{bmatrix}
s\mathbf{R} & \mathbf{t} \\
\mathbf{0}^\top & 1
\end{bmatrix}.
\end{equation}

To apply this transformation at the voxel level, the CT volume was resampled into the US image space as follows. Let $x_u \in \mathbb{R}^3$ is a voxel coordinate of the US volume expressed in homogeneous coordinates as $(x_u, 1)^T$. This coordinate is first converted to the world coordinate system using the affine matrix of the US volume $\mathbf{A}_{\text{US}}$. The corresponding world coordinate of CT volume is determined by applying the transformation $\mathbf{T}_{\text{US}\rightarrow\text{CT}}$. Finally, the inverse affine matrix of the CT volume $\mathbf{A}_{\text{CT}}^{-1}$ is applied to convert the world coordinates into the corresponding voxel coordinate of the CT. 

For efficient composition of these transformations, the single resampling matrix $\mathbf{M}$ is computed.
\begin{equation}
\mathbf{M} = \mathbf{A}_{\text{CT}}^{-1} \, \mathbf{T}_{\text{US}\rightarrow\text{CT}} \, \mathbf{A}_{\text{US}},
\end{equation}
 Using this transformation, CT intensity volumes were resampled using trilinear interpolation, while CT segmentation masks were resampled using nearest-neighbor interpolation to preserve discrete label values.

\subsection{Region-of-Interest Cropping}

After registration, the aligned CT and US volumes were spatially cropped to focus on the kidney region and remove irrelevant background. Cropping also reduces memory usage and improves learning efficiency by concentrating on anatomically meaningful regions.

The cropping region was determined using the US kidney segmentation mask, denoted by $\mathbf{M}_{\text{US}}$. The axis-aligned bounding box enclosing the kidney was computed as
\begin{equation}
\mathbf{b}_{\min} = \min \{ \mathbf{i} \mid \mathbf{M}_{\text{US}}(\mathbf{i}) > 0 \}, \quad
\mathbf{b}_{\max} = \max \{ \mathbf{i} \mid \mathbf{M}_{\text{US}}(\mathbf{i}) > 0 \}.
\end{equation}
A fixed margin of $5$ voxels was added in all spatial directions to retain surrounding anatomical context. The same bounding box was then applied to the registered CT image, CT mask, US image, and US mask, ensuring voxel-wise correspondence between both the modalities.

\subsection{Field of view masking}
\begin{figure}[h!]
    \centering
    \includegraphics[width=0.9\linewidth]{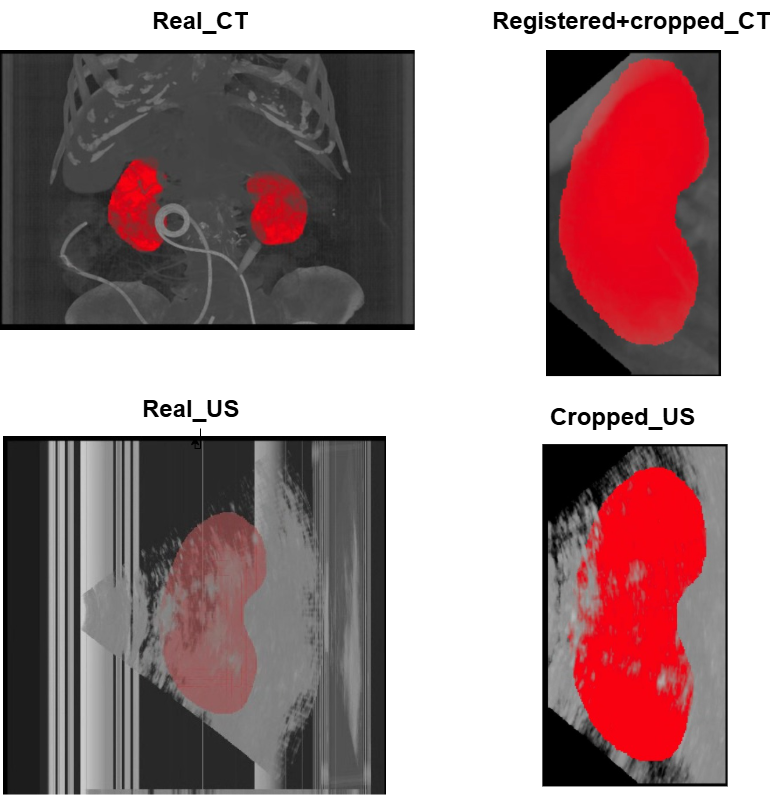}    
    \caption{Example of preprocessing for paired US–CT data. Real CT volumes are registered and cropped to the target anatomical region, while real US volumes are cropped accordingly to obtain spatially aligned training pairs.}
    \label{fig:registered}
    \vspace*{1mm}
\end{figure}

The registered CT volume is masked to match the field of view of the cropped US volume. In Fig.~\ref{fig:registered}, Cropped\_US illustrates the field of view of the US volume after cropping to the bounding box, which is inherently geometry-dependent.

The US image exhibits a pyramidal(frustum-shaped) field of view which is a characteristic of the probe based US acquisition. Only tissue regions that lie within this acoustic sector are reconstructed in the image. The regions outside this sector appear black. In constrast, CT images have full body field of view. In order to match the field of view of both the volumes, the registered, cropped CT is masked by the field of view mask of the US volume. The field of view mask $Mask_{FOV}$ of the US volume $US$ is computed by thresholding the US volume to zero. 

\begin{equation}
    Mask_{FOV}=
\begin{cases}
 1, & US>0 \\
0, & \text{otherwise}
\end{cases}
\end{equation}

The flood fill algorithm is applied to the $Maks_{FOV}$ to fill any possible voids in the binary mask. The $Mask_{FOV}$ is multiplied to the registered, cropped CT for field of view masking. Fig \ref{fig:registered}, Registered+Cropped\_CT shows the regions outside the field of view of Cropped\_US in black color.

\section{Methodology}
\label{sec:methodology}

In this work, we formulate a 3D UD-pCT generation framework as a supervised paired cross-modality image translation problem. Let a volumetric US scan be denoted by $\mathbf{U} \in \mathbb{R}^{1 \times D \times H \times W}$ and its corresponding CT volume by $\mathbf{C} \in \mathbb{R}^{1 \times D \times H \times W}$, where $D$, $H$, and $W$ represent the depth, height, and width of the 3D volumes, respectively. The objective is to learn a mapping function
\begin{equation}
\mathcal{G}: \mathbf{U} \mapsto \hat{\mathbf{C}},
\end{equation}
where $\hat{\mathbf{C}}$ denotes the predicted UD-pCT volume.

We propose a transformer-augmented residual U-Net framework for volumetric US-to-CT synthesis. The model is specifically designed to address key challenges associated with US imaging, including speckle noise, low contrast, and complex anatomical variability, while simultaneously preserving global anatomical consistency across the 3D volume.

\begin{figure*}[t]
  \centering
    \vspace*{-1mm} 
    \includegraphics[width=\textwidth,
    trim=0mm 220mm 0mm 0mm,
    clip]{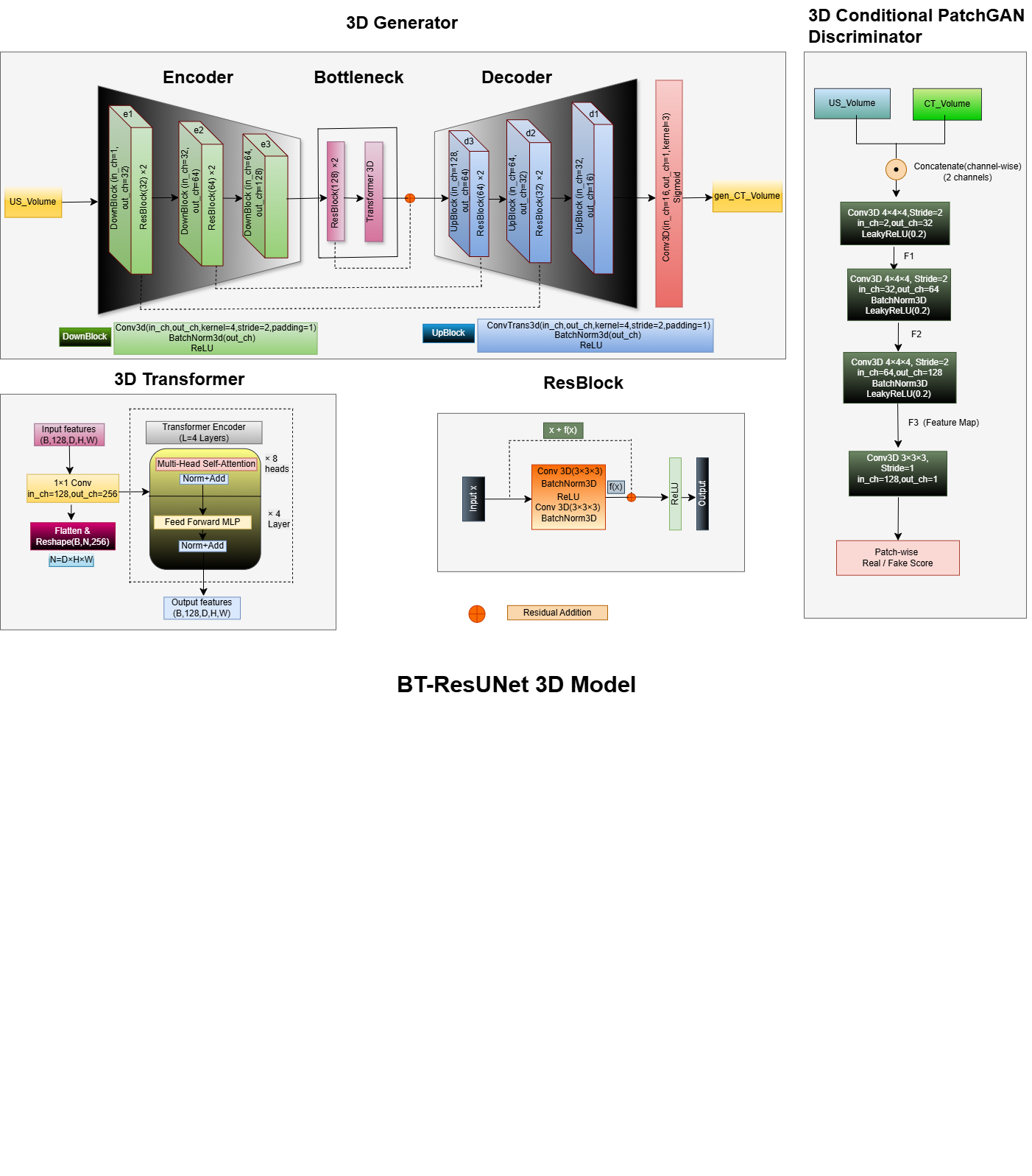}
    \caption{Overview of the proposed BT-ResUNet3D architecture. The generator consists of a 3D ResUNet encoder–decoder with residual blocks at each scale. A bottleneck transformer is introduced to model global 3D contextual relationships, and skip connections preserve fine-grained spatial information. The discriminator follows a 3D Conditional PatchGAN design for volumetric image-to-image translation. }
\label{fig:model}
\vspace*{-1mm} 
\end{figure*}

The proposed framework consists of three main components: (i) a 3D generator $\mathcal{G}$ implemented as a residual encoder--decoder network with a transformer bottleneck, (ii) a 3D PatchGAN discriminator $\mathcal{D}$ that enforces local volumetric realism through adversarial supervision, and (iii) a composite loss function that jointly enforces voxel-level accuracy and perceptual realism.
 An overview of the model architecture is provided in Fig.~\ref{fig:model}.

\subsection{Generator Architecture}
\label{subsec:generator}
The generator $\mathcal{G}$ follows a hybrid convolutional--transformer architecture that integrates local feature learning with global context modeling. Given an input US volume $\mathbf{U} \in \mathbb{R}^{1 \times D \times H \times W}$, the generator produces a synthesized pseudo-CT volume $\hat{\mathbf{C}} = \mathcal{G}(\mathbf{U})$ of the same spatial dimensions.

The overall architecture is designed to jointly capture:
\begin{itemize}
    \item Local texture and fine-grained anatomical details through deep residual 3D convolutional blocks,
    \item Hierarchical multi-scale representations via an encoder--decoder structure with skip connections, and
    \item Long-range volumetric dependencies using a transformer-based bottleneck for global context modeling.
\end{itemize}
\subsubsection{Residual 3D CNN Encoder--Decoder Backbone}

The generator backbone adopts a residual 3D convolutional encoder--decoder architecture inspired by the U-Net design. The encoder progressively extracts hierarchical feature representations from the input US volume while reducing spatial resolution. At each encoding stage, a downsampling operation is followed by multiple residual blocks to enhance feature expressiveness and stabilize optimization. Formally, the encoding operation at level $l$ is defined as:
\begin{equation}
\mathbf{E}_{l+1} = \mathcal{H}\big(\mathbf{E}_l; \phi_l\big) + \mathcal{F}\big[\mathcal{H}(\mathbf{E}_l; \phi_l); \theta_l\big], \quad l = 0, \dots, L-1,
\end{equation}
where $\mathbf{E}_0 = \mathbf{U}$, $\mathcal{F}(\cdot)$ denotes a residual function composed of two consecutive 3D convolutional layers with batch normalization and ReLU activation, $\theta_l$ represents the learnable parameters in $\mathcal{F}(\cdot)$ at encoder level $l$, $\mathcal{H}\big(.\big)$ denotes a encoder function consisting a 3D convolutional layer with batch normalization, ReLU activation function, stride 2, kernel size 4 and padding 1 and $\phi_l$ denotes the learnable parameters in the encoder function at level $l$.

Residual connections facilitate efficient gradient propagation and enable deeper feature learning, which is particularly important for capturing subtle anatomical structures in noisy US volumes.

The decoder mirrors the encoder structure at each layer and progressively restores the spatial resolution using transposed 3D convolutions. At each decoding level, the upsampled feature maps are fused with corresponding encoder features via skip connections:
\begin{equation}
\mathbf{D}_{l} = \mathcal{H'}\big(\mathbf{D}_{l+1}; \phi'_l\big) + \mathcal{F}\big[\mathcal{H'}(\mathbf{D}_{l+1}; \phi'_l); \theta'_l\big] + \mathbf{E_l}, \quad l = 0, \dots, L-1,
\end{equation}
where $\mathcal{H'}\big(.\big)$ denotes a decoder function composed of transposed 3D convolution with the same hyper-parameters as $\mathcal{H}\big(.\big)$ along with batch normalization followed by ReLU activation, $\phi'_l$ denotes the learnable parameters in the decoder function at level $l$ and $\theta'_l$ represents the learnable parameters in residual function $\mathcal{F} \big(.\big)$ at decoder level $l$.  This fusion preserves fine spatial details and ensures anatomical consistency between the input US and the synthesized pseudo-CT output.

\subsubsection{Transformer Bottleneck for Global Context Modeling}

While convolutional operations are effective at modeling local spatial patterns, their receptive field remains inherently limited. To capture long-range dependencies and enforce global anatomical coherence across the 3D volume, a transformer-based bottleneck is integrated at the deepest level of the encoder.

The encoder bottleneck feature map $\mathbf{B} \in \mathbb{R}^{C \times d \times h \times w}$ is first projected into a higher-dimensional embedding space using a $1 \times 1 \times 1$ convolution and then reshaped into a sequence of volumetric tokens:
\begin{equation}
\mathbf{T} = \text{Flatten}\big(\text{Conv}_{1\times1\times1}(\mathbf{B})\big) \in \mathbb{R}^{N \times C'},
\end{equation}
where $C' > C$ and $N = d \cdot h \cdot w$ denotes the total number of volumetric tokens.

The transformer encoder applies multi-head self-attention to model global interactions among all spatial locations:
\begin{equation}
\text{Attention}(\mathbf{Q}, \mathbf{K}, \mathbf{V}) =
\text{Softmax}\left(\frac{\mathbf{Q}\mathbf{K}^\top}{\sqrt{d_k}}\right)\mathbf{V},
\end{equation}
where $\mathbf{Q}$, $\mathbf{K}$, and $\mathbf{V}$ are learned linear projections of $\mathbf{T}$ and $d_k$ is dimensionality of keys $\mathbf{K}$(and queries $\mathbf{Q}$) for one attention head. The transformer output is reshaped back into volumetric form and combined with the original bottleneck features via a residual connection. This design enables global context to guide voxel-level reconstruction while preserving local structural information.

\subsubsection{Generator Output Layer}

The final pseudo-CT volume is produced by applying a $3 \times 3 \times 3$ convolution followed by a sigmoid activation:
\begin{equation}
\hat{\mathbf{C}} = \mathcal{G}(\mathbf{U}) = \sigma\big(\text{Conv3D}(\mathbf{D}_0)\big),
\end{equation}
where $\hat{\mathbf{C}} \in \mathbb{R}^{1 \times D \times H \times W}$ represents the synthesized pseudo-CT volume with voxel intensities normalized to the range $[0,1]$.

\subsection{3D Conditional PatchGAN Discriminator}
\label{subsec:discriminator}

The discriminator $\mathcal{D}$ is implemented as a \textbf{3D conditional PatchGAN} that evaluates the realism of local volumetric patches rather than assigning a single global score. Conditioning is achieved by concatenating the input US volume $\mathbf{U}$ with a corresponding CT volume, which may be either the real CT $\mathbf{C}$ or the generated pseudo-CT $\hat{\mathbf{C}}$, along the channel dimension:
\begin{equation}
\mathcal{D}(\mathbf{U}, \mathbf{C}), \quad
\mathcal{D}(\mathbf{U}, \hat{\mathbf{C}})
\in \mathbb{R}^{1 \times d' \times h' \times w'}.
\end{equation}

Each element of the discriminator output corresponds to the realism of a local 3D patch in the input volume. This patch-level discrimination encourages the generator to synthesize anatomically plausible textures and preserve sharp structural boundaries, which are critical for accurate ultrasound-to-CT translation. By operating in a fully volumetric manner, the proposed 3D conditional PatchGAN effectively enforces spatial consistency across depth, height, and width dimensions.

\subsection{Loss Functions}
\label{subsec:loss}

The generator is trained using a composite objective function that balances voxel-wise reconstruction fidelity and adversarial realism:
\begin{equation}
\mathcal{L}_{\mathcal{G}} =
\lambda_{\text{pix}} \mathcal{L}_{\text{pix}} +
\lambda_{\text{adv}} \mathcal{L}_{\text{adv}},
\end{equation}
where $\lambda_{\text{pix}}$ and $\lambda_{\text{adv}}$ control the relative contributions of the reconstruction and adversarial objectives, respectively.

\subsubsection{Pixel-wise Reconstruction Loss}

To enforce voxel-level correspondence between the generated pseudo-CT volume $\hat{\mathbf{C}}$ and the ground-truth CT volume $\mathbf{C}$, an $L_1$ loss is employed:
\begin{equation}
\mathcal{L}_{\text{pix}} =
\frac{1}{DHW}
\sum_{i,j,k}
\left|
\hat{\mathbf{C}}_{i,j,k}
-
\mathbf{C}_{i,j,k}
\right|.
\end{equation}

\subsubsection{Adversarial Loss}

To further enhance perceptual realism, an adversarial loss based on the 3D conditional PatchGAN discriminator is incorporated. Following the non-saturating GAN objective and using a binary cross-entropy loss with logits, the adversarial loss for the generator is defined as:
\begin{equation}
\mathcal{L}_{\text{adv}} =
\mathbb{E}_{\mathbf{U}}
\left[
\text{BCEWithLogits}
\big(
\mathcal{D}(\mathbf{U}, \hat{\mathbf{C}}),
\mathbf{1}
\big)
\right].
\end{equation}
where $\mathbb{E}_{\mathbf{U}}$ is expectation over $\mathbf{U}$.

By jointly optimizing the pixel-wise reconstruction loss and the adversarial objective, the proposed framework effectively balances anatomical fidelity and perceptual realism. The integration of residual learning facilitates stable deep feature extraction, while the transformer-based bottleneck captures long-range volumetric dependencies that conventional convolutional architectures fail to model. Together, these design choices enable the generation of structurally consistent and high-quality pseudo-CT volumes from noisy US inputs, outperforming conventional CNN-only and slice-based approaches.

\section{Experimental Setup}
\label{sec:experiments}

\subsection{Dataset and Splits}

Experiments were conducted on the TRUSTED dataset, which contains paired 3D transabdominal US and CT kidney volumes. After registration, intensity normalization, cropping around the kidney, and field of view masking, $N=59$ paired volumes were obtained. The dataset was split as follows to prevent data leakage at the patient level:

\begin{itemize}
    \item Training set: 47 cases
    \item Testing set: 12 cases
\end{itemize}

\subsection{Training Configuration}

All models were implemented in PyTorch and trained on a single GPU with mixed-precision. The Adam optimizer is used.

\textbf{Hyperparameters:}
\begin{itemize}
    \item Batch size: 1 (due to 3D memory constraints)
    \item Image size = $128\times128\times256$
    \item Epochs: 50
    \item Learning rate: $1 \times 10^{-4}$
    \item Adam betas: $(0.5, 0.999)$
    \item Base channels: 32
    \item Transformer projection dim: 256
    \item Transformer layers: 4
    \item Transformer heads: 8
    \item $\lambda_{\text{pix}} = 50, \lambda_{\text{adv}} = 5$
\end{itemize}

\section{Evaluation Metrics}
\label{sec:metrics}

To quantitatively evaluate the fidelity and structural consistency of the generated UD-pCT volumes, we employ two widely used image similarity metrics: Peak Signal-to-Noise Ratio (PSNR), and Structural Similarity Index Measure (SSIM). Let $\hat{\mathbf{C}} \in \mathbb{R}^{D \times H \times W}$ denote the generated 3D UD-pCT volume and $\mathbf{C} \in \mathbb{R}^{D \times H \times W}$ the corresponding ground-truth CT volume.

\subsubsection{Peak Signal-to-Noise Ratio (PSNR)}
PSNR is a logarithmic metric derived from Mean Squared Error(MSE) and measures reconstruction quality in decibels (dB):
\begin{equation}
\text{PSNR}(\hat{\mathbf{C}}, \mathbf{C}) = 10 \log_{10} 
\left( \frac{I_{\max}^2}{\text{MSE}(\hat{\mathbf{C}}, \mathbf{C})} \right),
\end{equation}
where $I_{\max}$ denotes the maximum possible voxel intensity value. In this work, all volumes are normalized to $[0,1]$, hence $I_{\max}=1$. Higher PSNR values indicate better reconstruction fidelity and lower noise. PSNR is commonly used in image synthesis tasks due to its interpretability and strong correlation with pixel-wise accuracy.

\subsubsection{Structural Similarity Index Measure (SSIM)}
SSIM evaluates perceptual similarity by jointly modeling luminance, contrast, and structural information. For two image patches $x$ and $y$, SSIM is defined as:
\begin{equation}
\text{SSIM}(x,y) =
\frac{(2\mu_x \mu_y + C_1)(2\sigma_{xy} + C_2)}
{(\mu_x^2 + \mu_y^2 + C_1)(\sigma_x^2 + \sigma_y^2 + C_2)},
\end{equation}
where $\mu_x, \mu_y$ are the mean intensities, $\sigma_x^2, \sigma_y^2$ are variances, and $\sigma_{xy}$ is the covariance between $x$ and $y$. Constants $C_1$ and $C_2$ stabilize the division.

SSIM is particularly important in medical imaging as it captures anatomical structure preservation and perceptual similarity beyond simple voxel-wise intensity differences.

\begin{figure*}[t]
    \centering
    \vspace*{-1mm} 
    \includegraphics[width=\textwidth]{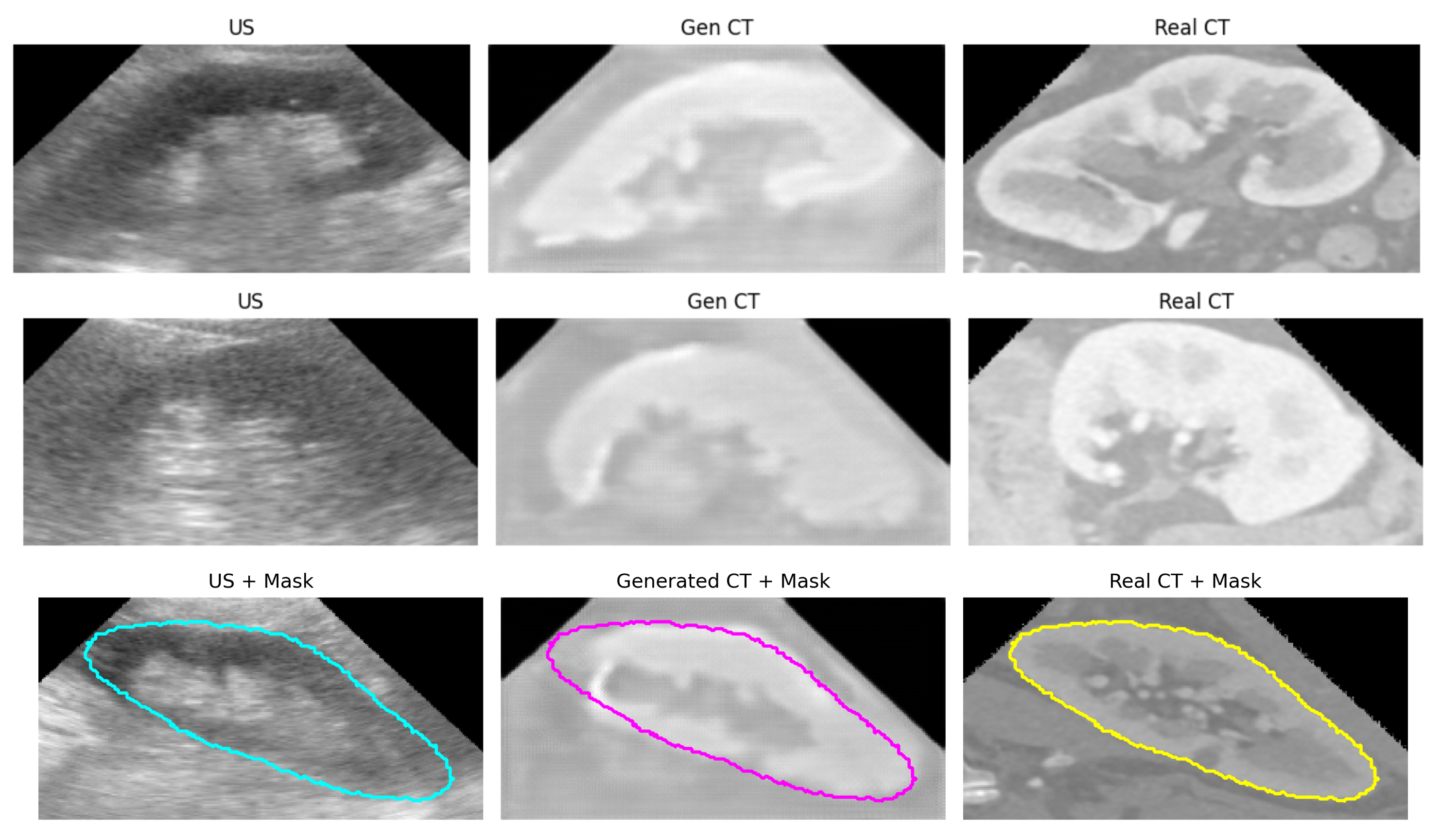}
    \caption{Qualitative comparison of US-to-CT synthesis results. From left to right: input US, generated UD-pCT, and CT. The first two rows show representative axial slices from different subjects. The third row visualizes kidney masks overlaid on US, UD-pCT, and CT images; the same mask derived from the real CT annotation is applied across all modalities for consistency.
}
\label{fig:qualitative_results}
\vspace*{-1mm} 
\end{figure*}
\begin{figure*}[t]
    \centering
    \vspace*{-1mm} 
    \includegraphics[width=\textwidth]{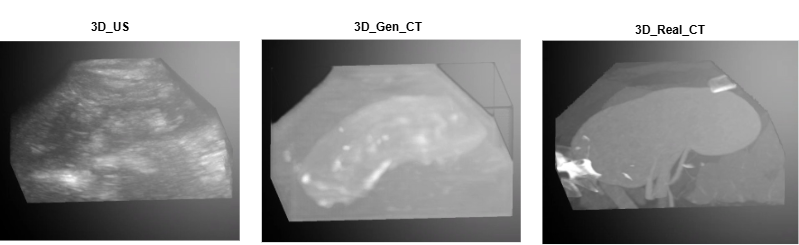}
    \caption{Visualization of UD-pCT synthesis. The input 3D US volume, the UD-pCT, and the CT volume are shown for comparison.
}
\label{fig:3D_qualitative_results}
\vspace*{-1mm} 
\end{figure*}
\section{Experimental Results and Analysis}
As no prior method specifically addresses paired 3D US-to-CT synthesis under the same objective, conventional convolutional and GAN-based models are treated as baselines, while diffusion-based approaches are considered state-of-the-art references. All models are trained on the same preprocessed dataset to ensure a fair comparison.
\begin{table}[htbp]
\centering
\caption{Quantitative comparison of baseline and state-of-the-art models for US-to-CT synthesis.}
\label{tab:baseline}
\begin{tabular}{lcc}
\toprule
\textbf{Model} & \textbf{PSNR (dB)} $\uparrow$ & \textbf{SSIM} $\uparrow$ \\
\midrule
MedDiffusion             & 6.279  & 0.0879 \\
3D Conditional Diffusion & 9.970  & 0.2752 \\
Pix2Pix                  & 22.574 & 0.6315 \\
3D ResUNet               & 22.862 & 0.6584\\
3D UNet                  & 22.187 & 0.6655 \\
\midrule
\textbf{Proposed Method} & \textbf{23.2593} & \textbf{0.7106} \\
\bottomrule
\end{tabular}
\end{table}

\subsection{Quantitative Results}

Table~\ref{tab:baseline} reports PSNR and SSIM comparisons. Baseline convolutional and GAN-based models achieve moderate performance, while diffusion-based methods perform poorly due to the large modality gap and limited paired data.  

An ablation study comparing standard 3D ResUNet and the proposed BT-ResUNet3D shows that adding the transformer-augmented bottleneck substantially improves both PSNR and SSIM, demonstrating the importance of modeling long-range volumetric dependencies. The proposed BT-ResUNet3D attains the highest scores (PSNR 23.26 dB, SSIM 0.71), reflecting improved structural fidelity and anatomical consistency.

\subsection{Qualitative Results}
Figures~\ref{fig:qualitative_results} and~\ref{fig:3D_qualitative_results} present representative synthesis results comparing the input US, the generated UD-pCT, and the corresponding real CT, shown as a 2D central axial slice and as 3D volumetric renderings. Despite the presence of speckle noise and low contrast in the US images, the proposed method successfully recovers coherent organ-level structures that are closely aligned with the underlying real CT anatomy.
Kidney masks derived from real CT annotations are overlaid to assess anatomical consistency. The synthesized UD-pCT closely follows the kidney boundaries and spatial extent observed in the real CT, demonstrating preserved organ-level geometry and accurate spatial alignment.

\section{Conclusion and Future Work}
In this work, we target the use of synthetic CT generation as a supportive tool for operator guidance during US acquisition, rather than as a replacement for diagnostically definitive CT imaging. By synthesizing pseudo-CT volumes directly from US data, the proposed approach provides operators with a CT-like anatomical reference that can enhance spatial awareness during scanning, support probe positioning and orientation, and help identify suboptimal acquisition in real time. This added anatomical context has the potential to reduce operator dependency and improve consistency in US examinations. 

By reducing reliance on unnecessary CT examinations, the proposed approach potentially lowers patient exposure to ionizing radiation and mitigating associated long-term health risks. Despite the limited size of the available paired dataset, our experimental results demonstrate that learning structurally consistent and anatomically meaningful CT-like representations from US is both feasible and promising.

Future research will focus on the acquisition of larger and more diverse paired US–CT datasets to further enhance synthesis fidelity and generalization. In addition, we plan to investigate advanced learning strategies aimed at reducing dependence on conventional CT scans, such as semi-supervised and weakly supervised training paradigms. We also intend to incorporate a robust intensity-based registration framework to improve alignment between US and CT volumes and reference anatomical spaces, thus further increasing model accuracy and reliability and reducing reliance on manually annotated landmarks. Given the widespread availability, low cost, and radiation-free nature of US imaging, the ability to generate CT-like volumetric representations from US could substantially decrease unnecessary CT utilization in clinical workflows. Overall, this study underscores the potential of US-based synthetic CT generation as a practical and safer alternative for improving guidance and efficiency in medical imaging applications.


\end{document}